%% file: root.tex
\documentclass[letterpaper, 10 pt, conference]{ieeeconf}  

\IEEEoverridecommandlockouts                              

\usepackage{graphicx}
\usepackage{fancyhdr}

\usepackage{macros_packages/packages}
\usepackage{hyperref}
\usepackage{booktabs}

\input{macros_packages/math}

\input{macros_packages/manip}

\input{macros_packages/utils}

\title{\LARGE \bf
Fast Estimation of Globally Optimal Independent Contact Regions \\
 for Robust Grasping and Manipulation}

\author{\authorblockN{Jonathan P. King\authorrefmark{1},
Harnoor Ahluwalia\authorrefmark{2},
Michael Zhang\authorrefmark{1}, 
and
Nancy S. Pollard\authorrefmark{1}}
\authorblockA{\authorrefmark{1}
The Robotics Institute, Carnegie Mellon University, Pittsburgh, PA 15224}
\authorblockA{\authorrefmark{2} Wakeland High School, Frisco, TX 75034}
\authorblockA{\authorrefmark{1}
Corresponding author: jking2@andrew.cmu.edu}
 }

\begin{document}

\maketitle

\begin{abstract}
This work presents a fast anytime algorithm for computing globally optimal independent contact regions (ICRs).   ICRs are regions such that one contact within each region enables a valid grasp.  Locations of ICRs can provide guidance for grasp and manipulation planning, learning, and policy transfer.   However, ICRs for modern applications have been little explored, in part due to the expense of computing them, as they have a search space  exponential in the number of contacts.   We present a divide and conquer algorithm based on incremental n-dimensional Delaunay triangulation that produces results with bounded suboptimality in times sufficient for real-time planning.  This paper presents the base algorithm for grasps where contacts lie within a plane.   Our experiments show substantial benefits over competing grasp quality metrics and speedups of 100X and more for competing approaches to computing ICRs.  We explore robustness of a policy guided by ICRs and outline a path to general 3D implementation.  Code will be released on publication to facilitate further development and applications.
\end{abstract}

\section{INTRODUCTION}

Independent contact regions have been recognized as a foundational tool for grasp planning~\cite{nguyen1986synthesis}.  Large ICRs suggest
grasps that can accommodate uncertainties and have geometries conducive to contact.  They have been employed
as stepping stones for in-hand manipulation~\cite{jeong2013hand,fontanals2014integrated} and it
 is interesting to consider how they may enhance
modern approaches to manipulation and grasping.    For example, ICRs could provide guidance for hand-object contact placement in language based interfaces~\cite{totsila2024words2contact,birr2024autogpt+}, they could improve our ability to interpret and retarget human demonstrations~\cite{liconti2024leveraging,saito2024apricot},  they could result in more physically informed grasp filters~\cite{burkhardt2024multi,feng2024dexgangrasp}, or enhance our ability to co-design hand shapes and product geometries~\cite{isakhani2024structural}.
However, ICR computation has been plagued by computational complexity.  Its search space is exponential in the number of contacts, and previous algorithms are practical only in special cases
(e.g.,~\cite{nguyen1988constructing,ponce1997computing}) or for locally optimal solutions (e.g.,~\cite{pollard2004closure,roa2009computation,krug2010efficient}).

\begin{figure}[t]    
    \centering
    \includegraphics[width=0.49\textwidth]{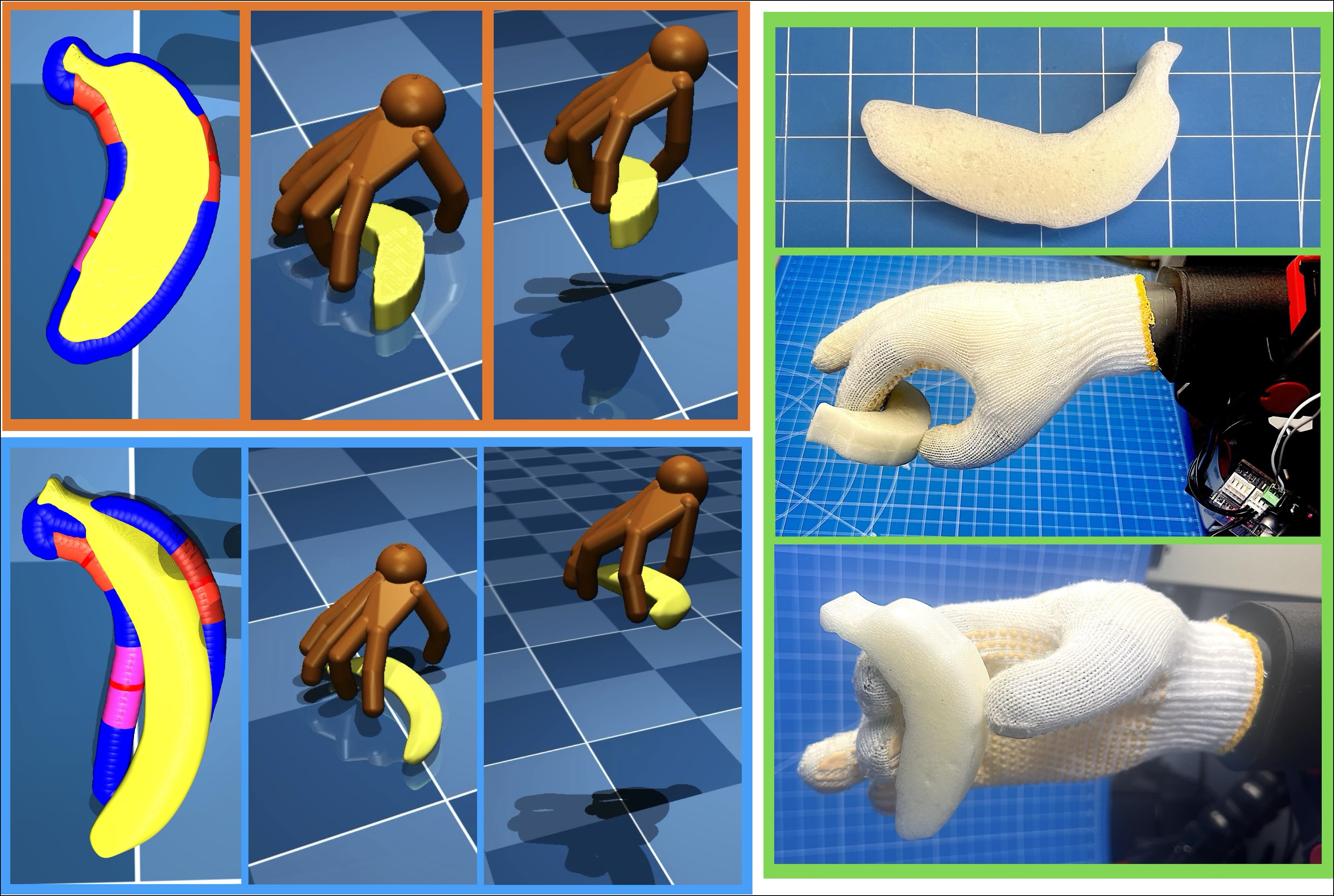}
    \vspace*{-0.25in}
    \caption{(Top) ICRs shown in orange and pink guide a simple control policy to lift an object.  (Bottom) Guidance provided by ICRs enables the policy to work even with errors in object size, location, and geometry. (Right) Results from policy transfer to an anthropomorphic hand.}
    \vspace*{-0.2in}
    \label{fig:teaser}
\end{figure}

This paper presents an algorithm for fast identification of globally optimal ICRs in high-contact scenarios.   The problem can be viewed as one of finding large empty regions in a high-dimensional volume.   This problem has been studied elsewhere (e.g.~\cite{toussaint1983computing}), and it has been observed that Delaunay Triangulation of occupied sites enables the largest empty circular region (hypersphere) to be identified easily. 
In our scenario, an occupied site corresponds to a grasp that is  undesirable.  In this paper we use the term ``invalid'' grasp, which could mean a grasp that would not achieve force closure, a grasp that contacts a forbidden region, or a grasp with measured quality below a user-specified threshold for some metric.  In our case, invalid grasps are not all known ahead of time, and so we develop an incremental Delaunay Triangulation algorithm that iteratively tests the largest empty region until the optimal (or a sufficiently large) set of ICRs is located.  Our algorithm guarantees convergence and produces results with bounded suboptimality.   Our results show nearly 100X speedup on average over a fast brute force implementation, and our algorithm enables for the first time $\epsilon$-optimal ICRs to be computed for grasps having as many as 7 contacts.   Furthermore, it can be used as an anytime algorithm, as it is able to return the current best solution with optimality bounds at any point.  ICRs returned by the algorithm can be used for policy guidance (Figure~\ref{fig:teaser}).  Simple policies built from ICRs show robustness to real-world uncertainties in object size, location, and geometry.


\begin{figure*}[ht!]    
    \centering
    \vspace*{-0.0in}
    \includegraphics[width=0.76\textwidth]{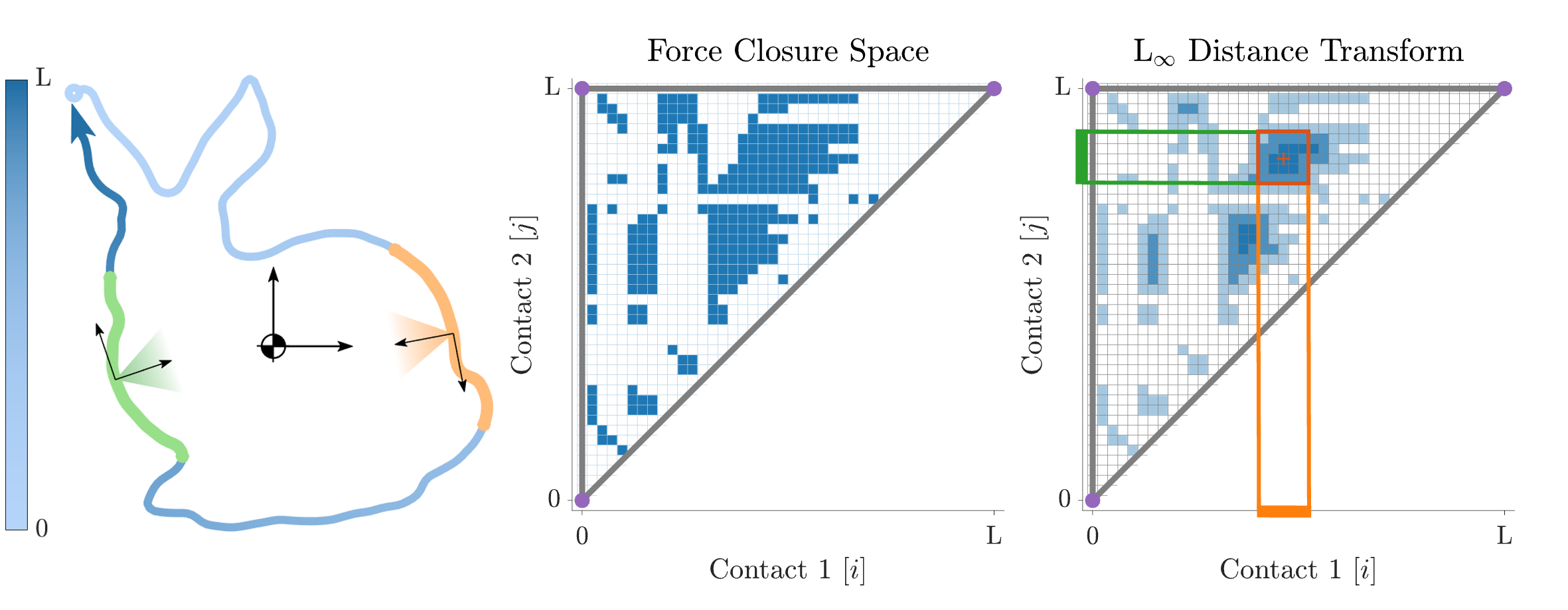}
    \includegraphics[width=0.22\textwidth]{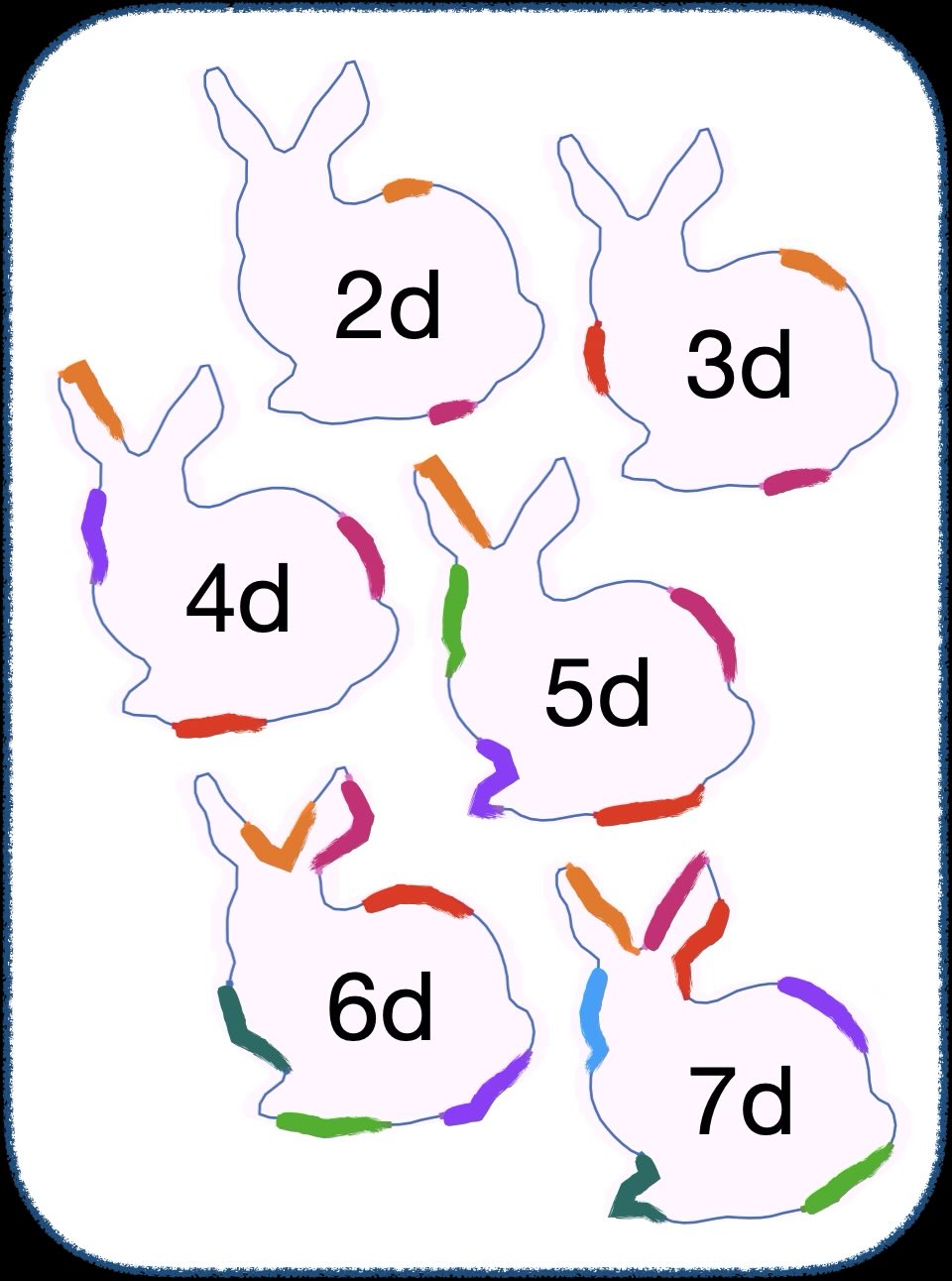}
    \caption{(Left) A shape defined by a curve parameterized on [0, L] with frictional contacts and contact regions. (Middle) The grasp configuration space as an order simplex with force closure grasps in blue. The $L_{\infty}$ Distance Transform computes the furthest interior point in the Chebyshev norm, corresponding to the largest square contained in the force closure space. The optimal ICR is shown in red made up of orange and green contact regions. (Right) Dimensionality of the order simplex goes up with number of contacts.  Our divide and conquer algorithm has produced the first result to our knowledge for epsilon-optimal grasps having greater than 4 contacts.  
    }
    \label{fig:bunny_fc_dt}
\end{figure*}

\section{Related Work}

Many algorithms for grasping and manipulation rely on measures of optimality or quality metrics~\cite{sahbani2012overview,bohg2013data,bicchi2000robotic}.
Historically, these methods have tended to produce optimal grasps as sets of precise target contact points.   As a result, it is easy to produce grasps that are impractical for a robot in the sense that an optimal grasp may be at some extreme, such as a small surface element or location that leads to instability in the face of position errors.   

Over time, partly because of difficulties such as these and partly due to trends in machine learning, quality metrics have evolved to include consideration of dynamics, uncertainty, and sensing and to evaluate the performance of policies based on real-world data and simulation~\cite{billard2019trends,tang2024deep}.  However, finding good policies remains challenging, and many impressive manipulation demonstrations benefit from physical understanding and control of contacts and forces~\cite{khadivar2023adaptive,yang2025multi}.  
 Metrics such as ICRs, if they can be computed quickly enough, can help to bring planning and policy based approaches together to handle constraints and uncertainties well while making good use of information available about the physics of an intended interaction.

Independent (Contact) Regions (ICRs) were introduced by Nguyen to provide robustness to grasps in the presence of fingertip position error and further developed by Ponce and Faverjon, with ICRs determined geometrically for special case 2, 3, and 4-contact grasps in the plane~\cite{nguyen1988constructing,ponce1995computing,ponce1993computing}.     Pollard computed ICRs for arbitrary numbers of contacts by growing regions from example grasps~\cite{Pollard1994,pollard2004closure}, although these ICRs were not globally optimal.  Cornella and Suarez proposed an algorithm for computing ICRs for n-contacts by working in the 2D object space~\cite{cornella2005fast}.   These results are also not globally optimal as only a subset of the force closure space is considered. There has additionally been work in finding ICRs on 3D objects~\cite{roa2009computation}, use of various contact models~\cite{charusta2012independent,roa2011influence}, integration of kinematic reachability~\cite{fontanals2014integrated}, and planning in-grasp manipulations such as finger gaiting~\cite{phoka2003regrasp}.

Most related to our contribution is work by Phoka and colleagues~\cite{phoka2012optimal}, who make the connection that for two-contact grasps, the axis-aligned rectangle in contact space that corresponds to any ICR (e.g., as shown in Figure~\ref{fig:bunny_fc_dt}) must be centered on
the $L_{\inf}$ Voronoi diagram constructed from the polygon bounding
the force closure regions~\cite{phoka2012optimal}.  The algorithm in~\cite{phoka2012optimal} is presented for grasps with two contacts.   Challenges to extending to more than two contacts include developing a stable software library for $L_{\infty}$ Voronoi diagrams (see~\cite{bukenberger2022constructing} for state of the art), and constructing volumes of higher-dimensional force closure regions -- a pre-processing step that is exponential in number of contacts, in contrast to the divide and conquer algorithm that we propose here.    We compare to their approach in Section~\ref{sec:results}.

There have been other algorithms focused on finding the largest empty axis aligned box in a point set (e.g., ~\cite{dumitrescu2013largest, lemley2016big}).   However, these algorithms assume that all invalid points are known.   The algorithm presented in this manuscript is adapted to the situation where invalid points are discovered incrementally.   We are motivated by approaches to finding the largest empty circle contained in a bounded region (e.g., ~\cite{toussaint1983computing}), and build on this work to develop an anytime algorithm with convergence guarantees when invalid points are not known in advance.


\section{Technical Implementation}

The brute force process for computing ICRs is illustrated in
Figure~\ref{fig:bunny_fc_dt} with a two-contact example.  For purposes of discussion, assume a 2D object with contact on the perimeter.   We will discuss how to lift those restrictions later in the paper.  The object
perimeter is discretized into $L$ intervals as shown in
Figure~\ref{fig:bunny_fc_dt}.  Contact 1 can be placed anywhere
along the perimeter.  We
assume that contacts are ordered and cannot overlap.  Contact 2 therefore must
be placed at a location with greater index than Contact 1, resulting in a simplex of possible grasps as shown in the middle of Figure~\ref{fig:bunny_fc_dt}.  Each grasp can be evaluated based on
any desired quality metric to determine whether it is ``valid'' or
``invalid.''  In the middle image of the figure, valid grasps based on the force closure property are
colored blue.  Once all possible grasps are classified, optimal ICRs
are found as the largest axis aligned box having all valid grasps and
contained within the simplex.  We can quickly identify the optimal
result for this example using the $L_{\infty}$ metric distance transform,
which is shown with shaded blue boxes in the right image of Figure~\ref{fig:bunny_fc_dt}.  The resulting optimal ICRs are highlighted, along with
corresponding contact regions on the bunny perimeter in orange and
green.   A more formal description follows.

\subsection{Grasp Configuration Space: The Order Simplex}

Formally, a contact lies on the parameterized object boundary $\partial B$ 
 according to $c \in [0, L]$. A grasp $\grasp$ consists of $d$ non-coincidental contacts.  We assume that contact regions cannot overlap, resulting in a grasp configuration space $C_d$ written
\begin{equation}
    C_d \ceq \left([0,L]^d \setminus \Delta^d_{[0,L]}\right)/\text{Sym}(d)
\end{equation}
where we begin with the 
$d$-fold Cartesian product of the contact manifold $[0,L]$ with itself, remove the fat diagonal $\Delta$, and then take the quotient space by the symmetric group $\text{Sym}(d)$, removing permutations in contact ordering. This is the space of unordered and unlabeled configurations of $d$ contacts. 
Geometrically, we can represent $C_d$ with an Order Simplex~\cite{grotzinger1984projections} by symmetrically dissecting the hypercube, reducing the volume by a factor of $d!$. An Order Simplex for $d$-contacts is written as
\begin{equation}
    \Delta^d\ceq\{t\in[0,L]^d: 0\leq t_1 \leq t_2 \leq \cdots \leq t_d \leq L\}
\end{equation}




\begin{algorithm}[ht]   
    \caption{Optimal ICR Algorithm}
    \label{alg:main}
    \DontPrintSemicolon

    \SetKwFunction{oSimplex}{OrderSimplex}
    \SetKwFunction{Delaunay}{DelaunayTriangulation}
    \SetKwFunction{Candidate}{BestCandidate}
    \SetKwFunction{Grow}{GrowHypercube}



    \KwData{A shape object $O$, number of contacts $d$}
    \KwResult{The optimal ICR $I^*$} 

    $\Delta^* \gets $ \oSimplex{$O, d$}\;   
    $DT \gets $ \Delaunay{$\Delta^*$}\;
    \tcc{Initialize ICR $I^*$, best candidate $C^*$}
    $I^* \gets \emptyset, C^* \gets $ \Candidate{$DT$}, $\epsilon \gets \infty$ 

    \tcc{while a better hypercube is possible}
    \While{$C^*.\mathtt{radius} > \max(I^*.\mathtt{radius}, r_{min})$}{
        \tcc{get new ICR $I$ and invalid grasp $G$}
        $I, G \gets $ \Grow{$C^*$} \\
        \tcc{if new best radius, store result}
        \If{$I.\mathtt{radius} > 
        I^*.\mathtt{radius}$}{
            $I^* \gets I$
        }
        \tcc{update $DT, C^*, \epsilon$}
        $DT.\mathtt{insert\_point}(G)$\;
        $C^*.isExplored = True$\;
        $C^* \gets $ \Candidate{$DT$}  \\
        \If{$I^*.\mathtt{radius} > 0$} {
            \tcc{get current fractional bound}
            $\epsilon \gets \frac{C^*.radius - I^*.radius}{I^*.radius}$
        }
    }
    \KwRet $I^*$
\end{algorithm}    

\subsection{Algorithm Setup}

\smallskip\noindent
{\bf Independent Contact Regions} are assumed to be disjoint continuous regions on the object boundary $\partial B$ such that any grasp $\grasp$ with contacts in the same ICR can be classified as a valid grasp.   Within the grasp configuration space $C_d$ an ICR is an axis aligned box that contains only valid grasps.

\begin{figure*}[ht!]    
    \centering
    \includegraphics[width=0.9\textwidth]{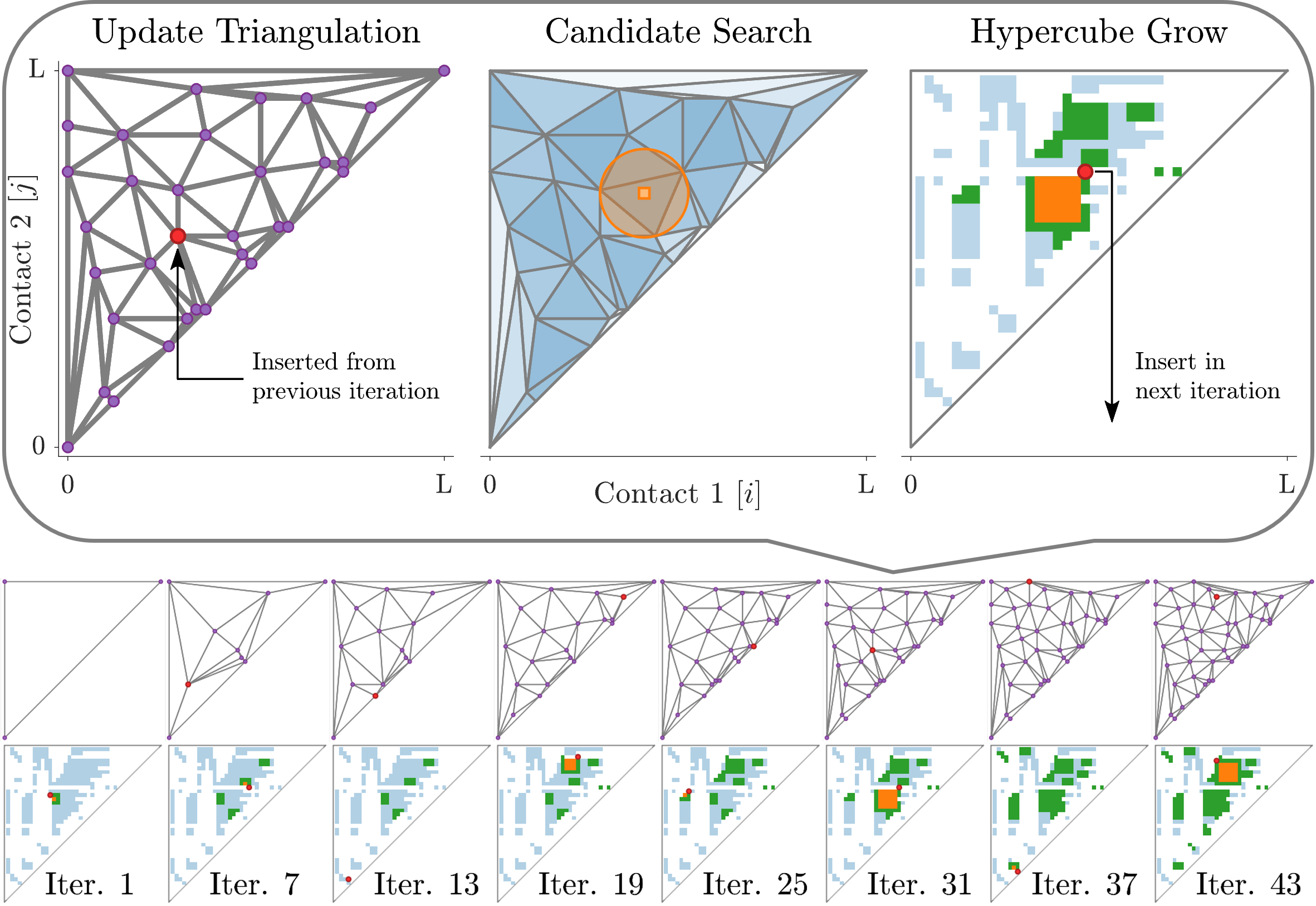}
    \caption{Illustration of iterative algorithm for computing Independent Contact Regions. (Top) A single iteration consists of identifying the best next candidate, i.e., the largest current empty region and attempting to grow a hypercube within that region.   Once an invalid grasp is found, the triangulation is updated and the next iteration begins.  Color coding is as follows:  (left) purple vertices are invalid grasps; (middle) blue shading in triangles is darker for higher candidate scores;  the orange circle is the circumsphere of the highest scoring candidate;  (right) light blue are valid grasps shown for visualization only;  green are valid grasps already discovered by the algorithm;  orange is the largest hypercube that could be grown at the current candidate center (orange circle);  red is the first invalid grasp found during hypercube growing. (Bottom) Snapshots at different stages in the process.}
    \label{fig:iter_illustration}
\end{figure*}

\smallskip\noindent {\bf Algorithm Concept}: Consider a set of valid and invalid grasps that have been identified within $C_d$. For the invalid grasp set
$\overline{V}$, consider it's Delaunay Triangulation $DT =
\textsc{DelaunayTriangulation}(\overline{V})$. The Delaunay property
guarantees an empty circumsphere for every simplex, $\sigma \in DT$,
in the triangulation.  The algorithm that then comes to mind is to
search within the largest empty circumsphere until an invalid grasp is
found then update the triangulation, repeating the process until there
can be no ICRs remaining that are larger than the largest already found within some bound.
This is the basic idea of our approach (Figure~\ref{fig:iter_illustration}).

\begin{figure}[h]     
    \centering
    \includegraphics[width=0.6\columnwidth]{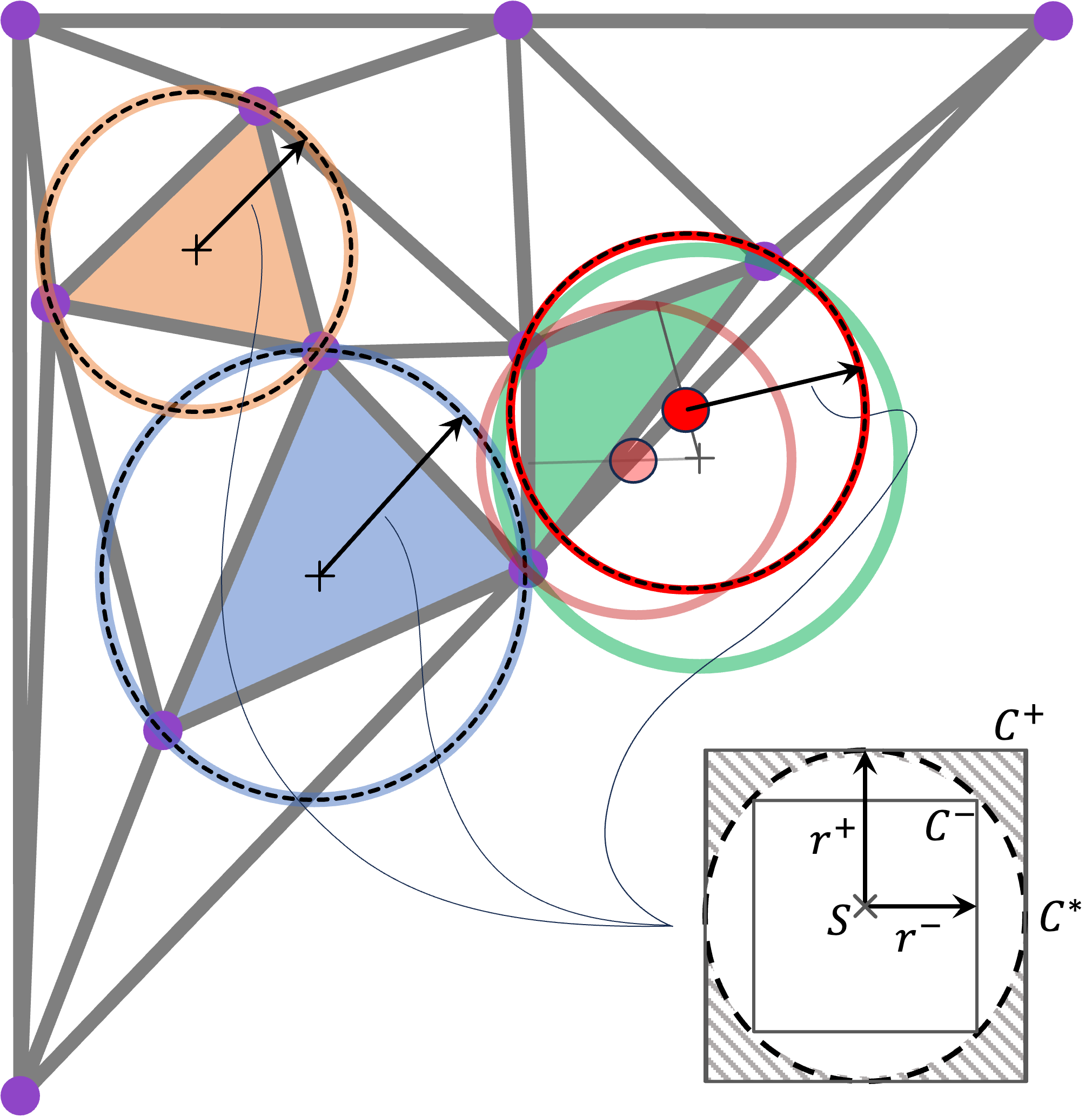}
    \caption{Each simplex in the Delaunay Triangulation has a candidate score. When the circumcenter is in the interior of the Order Simplex (orange and blue), the score is the circumradius.   Circumcenters outside the order simplex will undergo further processing (Section~\ref{sec:practicalities}).
    (Inset) For a given hypersphere $C^*$ with center $S$, the smallest hypercube $C^-$ centered at $S$ is the inscribed hypercube with half-edge length $r^-$ and the largest hypercube $C^+$ centered at $S$ is the circumscribed hypercube with half-edge length $r^+$.}
    \label{fig:candidate_score}
\end{figure}         

\smallskip\noindent {\bf Bounds}:
Delaunay Triangulation partitions the Order Simplex into smaller simplices, each having an enclosing circumsphere which has either already been explored or is empty.  Enclosing circumspheres can be used to bound the largest possible empty axis
aligned boxes, and hence the largest possible ICRs.  For any small and positive $\epsilon_r$, bounds for a hypercube placed at the center $S$ of an empty circumsphere can be expressed as (Figure~\ref{fig:candidate_score} Inset):
\begin{equation}
(r^{-}-\epsilon_r) < r_b < r^{+}
\label{eq:bound1}
\end{equation}
\begin{equation}
r^{-} = r^{+}/\sqrt{d}
\label{eq:bound2}
\end{equation}
or alternatively:
\begin{equation}
\epsilon = \sqrt{d} - 1
\label{eq:epsilon1}
\end{equation}
\begin{equation}
   r^- * (1 + \epsilon) = r^+ 
   \label{eq:epsilon2}
\end{equation}
where $r_b$ is the radius of the largest possible empty axis aligned box, expressed as its half-edge length, and $r^{+}$ is the radius of the largest empty circumsphere.  

\smallskip\noindent
{\bf Algorithm Iteration (Algorithm~\ref{alg:main})}:   The baseline algorithm works as follows.   In each iteration, the simplex with the highest candidate score is found (the simplex having the largest empty circumsphere).   A hypercube is grown from the center of the largest empty circumsphere until an invalid grasp is encountered.   
After hypercube growth, the best ICR is updated, and the invalid grasp is inserted into the Delaunay Triangulation $DT$.
There are several cases for how the algorithm may terminate or proceed. 
 In {\bf Case 1}, the radius of the best candidate $(C^{*}.radius)$ is less than or equal to the radius of the largest empty axis aligned hypercube found $(I^{*}.radius)$.   In this case the algorithm will find  no better set of Independent Contact Regions by proceeding than has already been found and the algorithm terminates.   In {\bf Case 2}, an invalid grasp $G$ is found which is already contained in the point set of $DT$.   For the base algorithm in Algorithm~\ref{alg:main}, we mark candidate $C^{*}$ as explored and proceed to the next iteration.  In {\bf Case 3} an invalid grasp $G$ is found that is not contained in $DT$.   This grasp is added to $DT$ and $DT$ is updated.   $C^*$ is marked as explored.  The next best candidate is identified, and the algorithm proceeds.



\subsection{Completeness and Convergence}

\smallskip\noindent
{\bf Lemma}: {\it An iteration of the algorithm will find a point which causes subdivision, mark a simplex as explored, or both.}

\smallskip\noindent
{\it Proof}. Let $\sigma$ be the simplex in $DT$ whose circumradius is maximal.   There are two cases which must be considered.

\smallskip\noindent
{\bf Case 1}: While growing an axis-aligned cube $C$ centered on $s = circumsphere(\sigma)$, an invalid point $b \subset s$ is encountered. $b$ is inserted into $DT$ which causes $\sigma$ to be subdivided. The algorithm continues on with the next iteration.

\smallskip\noindent
{\bf Case 2}: While growing $C$, an invalid point $b \not\subset s$ is encountered, meaning that $C^- \subseteq C \subsetneq C^+$. Inserting $b$ into $DT$ will not sub-divide $s$, although it may subdivide some other circumsphere.   Although $s$ may remain as part of the triangulation, it is marked as explored and will not be revisited.   $\square$

\smallskip\noindent
{\bf Proposition}: {\it The algorithm will terminate with no grasp larger than $r_{min}$ or with ICR having radius within $\frac{1}{\sqrt{d}}$ of optimal.}

\smallskip\noindent
{\it Proof outline.}
Following the preceding lemma, every iteration of the algorithm will reduce the number of candidates at the current radius or reduce the size of the optimal candidate radius (i.e., the size of the largest empty circumsphere cannot increase by adding invalid points).   This process will continue until one of the following things happens:

\smallskip\noindent
{\bf Case 1:}  The best candidate radius reaches $r_{min}$ and no grasp of size larger than $r_{min}$ has been found.   We can conclude that a reasonably sized ICR is not available.

\smallskip\noindent
{\bf Case 2:}  Let $r_{max}$ be the size of the largest hypercube found. In this case, the algorithm exits because the best candidate radius reaches a size less than or equal to $r_{max}$.  At this stage of the algorithm, the Delaunay Triangulation contains only the following:  (A) Unexplored simplices with circumspheres having radius less than or equal to $r_{max}$; (B) Already explored simplices with circumspheres which may be greater than $r_{max}.$
Choose the remaining circumsphere with the largest radius $r^+$.   The empty interior is the largest empty circle given the set of known invalid grasps (i.e., given the current Delaunay Triangulation).

\smallskip\noindent
{\bf Lemma.}  {\it There can be no empty axis aligned hypercube with radius greater than $r^+$.}

\smallskip\noindent
{\it Proof.}   Suppose there is one.   The inscribed circumsphere of this hypercube would be empty and would have radius greater than $r^+$.
However, the Delaunay property tells us that the largest possible empty hypersphere has a radius of $r^+$.  We have a contradiction. $\square$

\smallskip\noindent
{\bf Lemma.}  {\it There is a hypercube inside this circumsphere with radius $r^- - \epsilon_r < r_b < r^+$. }

\smallskip\noindent
{\it Proof.}   The simplex is explored by comments above.  The simplex was not subdivided.   By arguments outlined earlier in the manuscript, when the simplex was explored, any invalid grasp found must have been outside $C^*$, leading to the given bounds.  $\square$

\smallskip\noindent
From the Lemmas above and expressions in Equations~\ref{eq:epsilon1} and~\ref{eq:epsilon2}, if we terminate in Case 2, we must have found a solution such that $r_{max} \geq \frac{r^+}{\sqrt{d}}$, an $\epsilon$-optimal solution. 
$\square$

\subsection{Practical Considerations}
  \label{sec:practicalities}

\noindent{\it External Circumspheres:}
Observe that simplices defined by points inside the order simplex may have circumspheres with extent outside the order simplex (e.g., the green circle in Figure~\ref{fig:candidate_score}). In practice, this
situation occurs often.  Small simplices near the boundary of the order
simplex can masquerade as large empty spaces, when they could not
 contain large size ICRs under our problem
definition.
We can use results from prior investigation of largest empty circle with boundary (e.g.,~\cite{toussaint1983computing}) to address this challenge.   Given that the region center must be contained within the Order Simplex, we know that the largest empty hypersphere will be centered either at an interior circumcenter of the Delaunay Triangulation (a Voronoi point) or at the intersection of Voronoi elements with the convex hull of the Order Simplex~\cite{toussaint1983computing}.  In our case a grasp on the convex hull of the Order Simplex is not valid.   We can use this result to select potential points to add to the triangulation in order to find the best next candidate to explore.   In practice, we walk the Voronoi edges to find any intersections of those Voronoi edges with the Order Simplex convex hull, compute the radii of  maximal circumcenters at those points, and store the largest as the candidate score $r^{+}$, along with the invalid point that should be added if that circumcenter becomes the best candidate to be investigated. Figure~\ref{fig:candidate_score} shows an example.     It can be shown that this practice guarantees monotonic convergence of candidate scores for two contact grasps following the arguments in~\cite{toussaint1983computing}.  For grasps having greater than two contacts, it provides a significant practical speedup to the base algorithm.

\medskip\noindent
{\it Score Adjustment:}   It is a rare case where we can grow a hypercube to nearly the full radius $r^{+}$.   Even if the interior region of the circumsphere consists entirely of valid grasps, we will encounter one of the simplex points that define the circumsphere before we reach $r^+$.   We can easily compute the maximum radius of the hypercube allowed by the simplex points.  This radius can be used as a more accurate candidate score and results in more direct convergence.

\medskip\noindent
{\it Early Exit.} Algorithm~\ref{alg:main} guarantees we will find a hypercube radius within $\frac{1}{\sqrt{d}}$ of optimal.  We can choose to exit once we have reached that bound.   We do so by monitoring the value of $\epsilon$ in Algorithm~\ref{alg:main}.   Following Equations~\ref{eq:epsilon1} and~\ref{eq:epsilon2}, once we reach an $\epsilon$ value of $(\sqrt{d} - 1)$, we have reached the bounds guaranteed by our algorithm and we may choose to exit at that point.   Practically, a user can set any epsilon they like or exit after a fixed time period.  The current optimal result and epsilon bound are available at the end of each iteration.

\section{Results}
  \label{sec:results}

Experiments are detailed below.\footnote{All experiments were conducted on a MacBook Pro Apple M2 Max with 32GB of RAM and the software was implemented in Python 3.12.}

\medskip\noindent
{\bf State of the Art Comparison: $L_\infty$-Voronoi Diagrams.}
We begin with a state of the art comparison vs. the closest competing research~\cite{phoka2012optimal}.   Phoka and his colleagues identify the largest ICR by first constructing the $L_{\infty}$ Voronoi Diagram of the polygonal boundary representation of the force closure regions.   Their algorithm was implemented for grasps having 2 contacts, and we compare against their published examples.  Our algorithm is sampling-based, and we follow the comparison approach recommended in their paper of using 500 samples for each object (see~\cite{phoka2012optimal}).

\begin{table*}[t]   
    \centering
    \caption{Comparison with Phoka et al.~\cite{phoka2012optimal}.   Figures a through d, resolutions $n$ (\cite{phoka2012optimal}/ours) and friction cones $\mu = 10^{\circ}, 15^{\circ}, and \; 20^{\circ}$.   Runtimes are:  time of first valid solution ($t_1$), time of optimal solution ($t_o$), time to prove solution is optimal ($t_i$), time reported in Phoka et al.~\cite{phoka2012optimal} ($t_p$).  Speedups compared to reported results (e.g., $t_1/t_p$) are shown in parentheses.}
    \begin{tabular}{ll|llll|llll|llll}
        \toprule
        {}   &  {} & \multicolumn{4}{c}{$10^{\circ}$}       & \multicolumn{4}{c}{$15^{\circ}$}         & \multicolumn{4}{c}{$20^{\circ}$}      \\
        Fig. &  n  & t$_1$ (ms) & t$_o$ (s)  & t$_i$ (s)  & t$_p$ (s)     & t$_1$ (ms) & t$_o$ (s)  & t$_i$ (s)    & t$_p$ (s)      & t$_1$ (ms) & t$_o$ (s)  & t$_i$ (s)   & t$_p$ (s)    \\
        \midrule
        a    & 62/500  &   {\bf 28} (36)   &   1.1 (1) &  1.1 (1) & 1.02 &   {\bf 10} (117)   &   {\bf 0.22} (5)  &   {\bf 0.23} (5) &  1.17   &   {\bf 0.4} (4K)   &   {\bf 0.13} (12) &  {\bf 0.13} (12) & 1.61 \\
        b    & 100/500 &   {\bf 50} (19)   &   {\bf 0.47} (2) &   0.53 (2) & 0.95 &   {\bf 10} (253)   &   {\bf 0.25} (10)  &   {\bf 0.25} (10) & 2.53  &   {\bf 8} (339)   &   {\bf 0.15} (19) &   {\bf 0.16} (17) & 2.78\\
        c    & 200/500 &   {\bf 8} (338)   &   {\bf 0.25} (11) &   3.42 (1) & 2.70 &   {\bf 3} (1K)   &   {\bf 0.31} (14)  &   3.37 (1) & 4.32  &   {\bf 3} (2K)   &   {\bf 0.41} (14) &   {\bf 0.41} (14) & 5.67\\
        d    & 300/500 &   {\bf 2} (3K)   &   2.05 (3)  &   3.42 (2) & 5.50  &   {\bf 2} (4K)   &   0.83 (9)   &  3.37 (2) & 7.80   &   {\bf 2} (5K)   &   2.89 (4) &  3.38 (3) & 10.36 \\
        \bottomrule
    \end{tabular}
    \label{tab:phoka}
\end{table*}

\begin{figure*}[ht] 
    \centering
    \includegraphics[width=0.28\columnwidth]{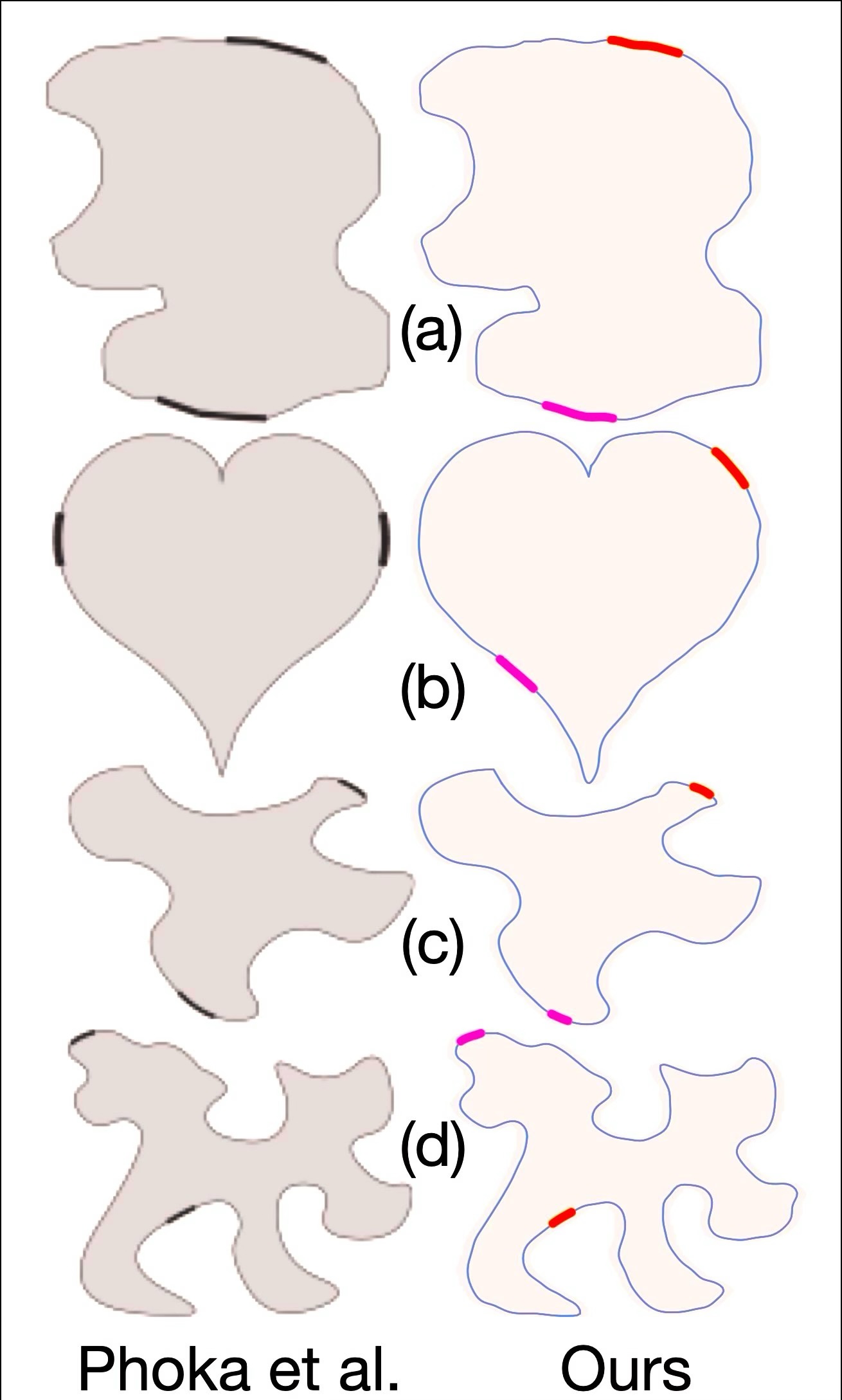}
    \includegraphics[width=0.65\columnwidth]{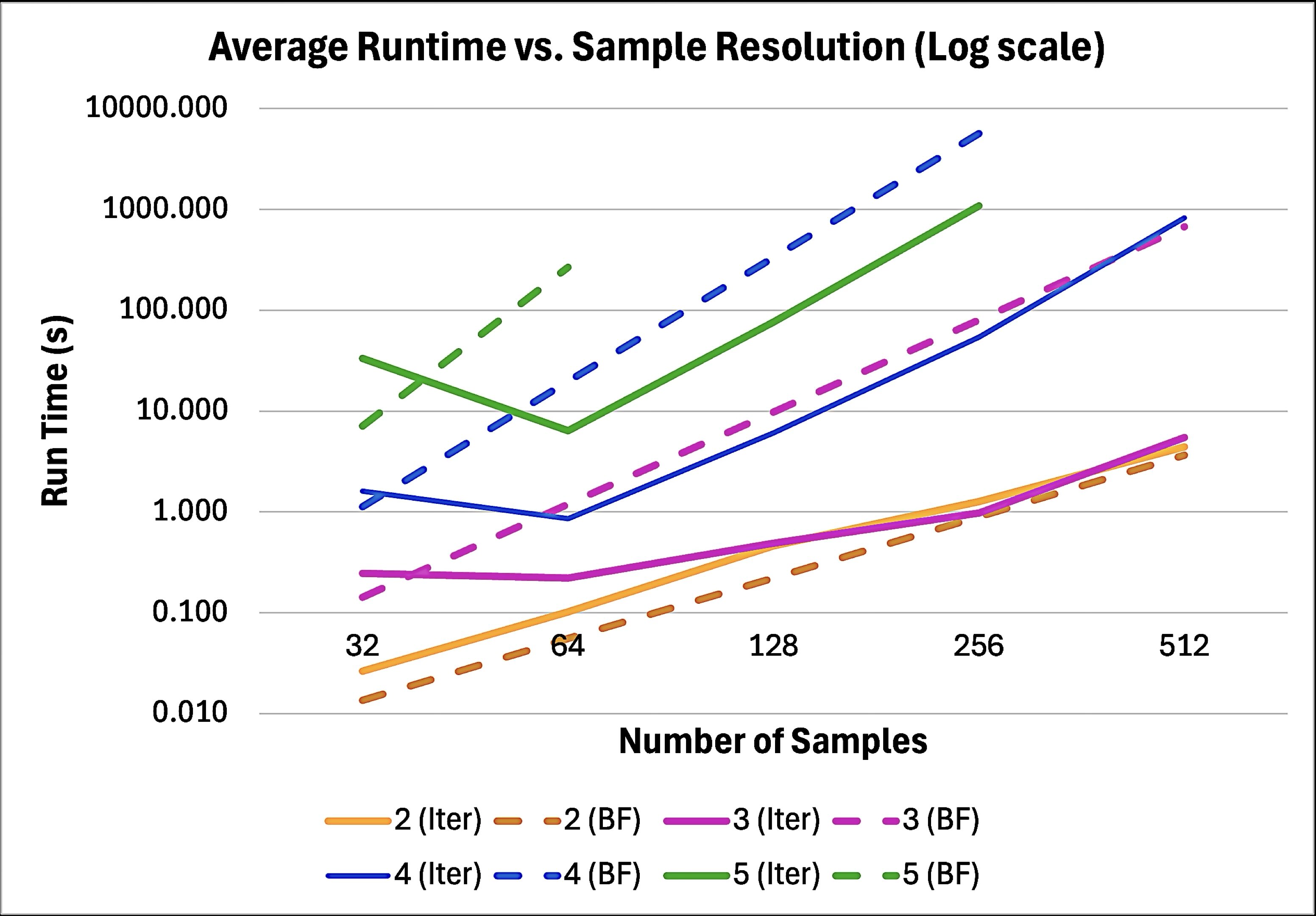}
    \hspace*{0.0in}
    \includegraphics[width=0.45\columnwidth]{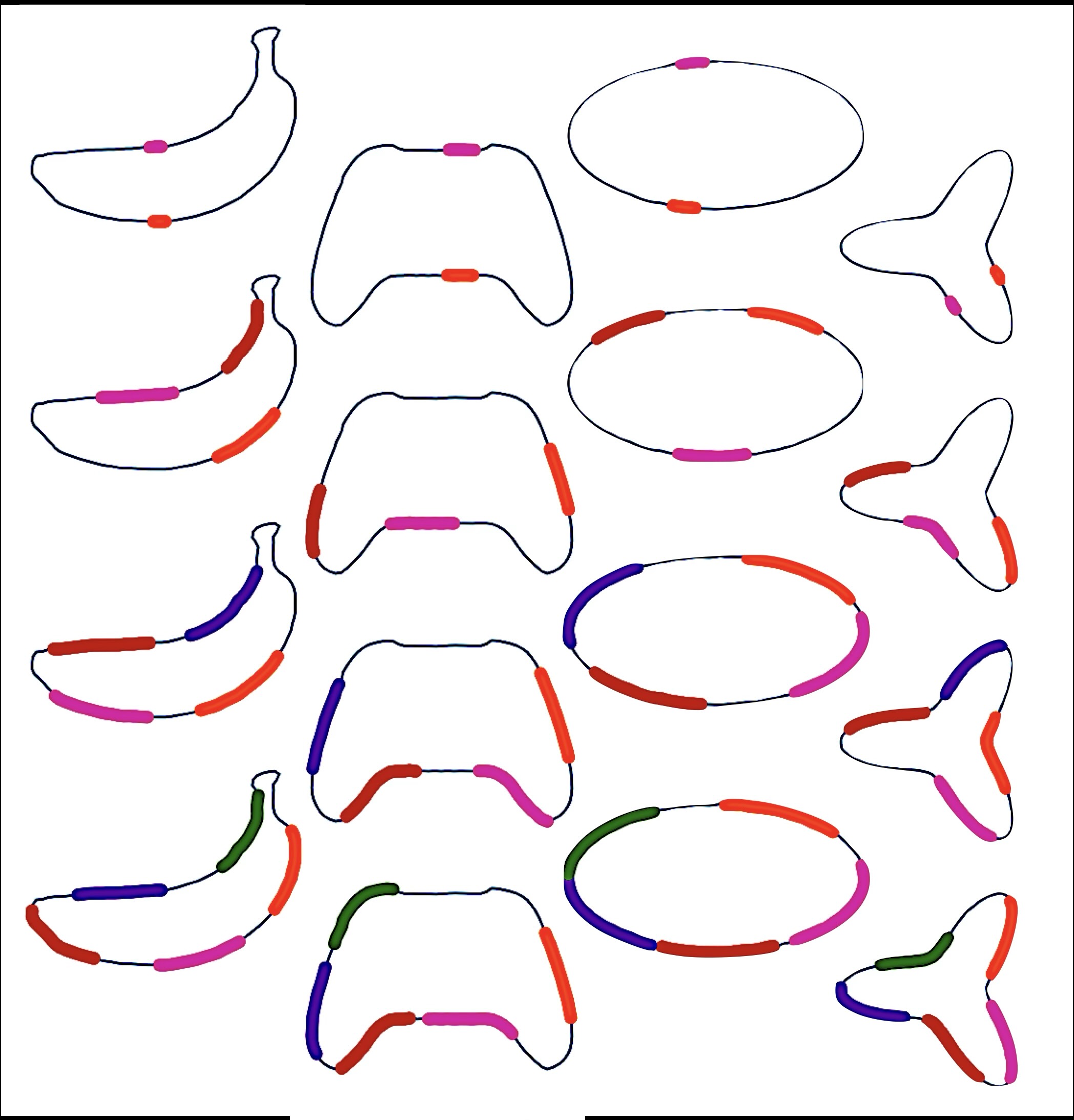}
        \hspace*{0.0in}
        \includegraphics[width=0.6\columnwidth]{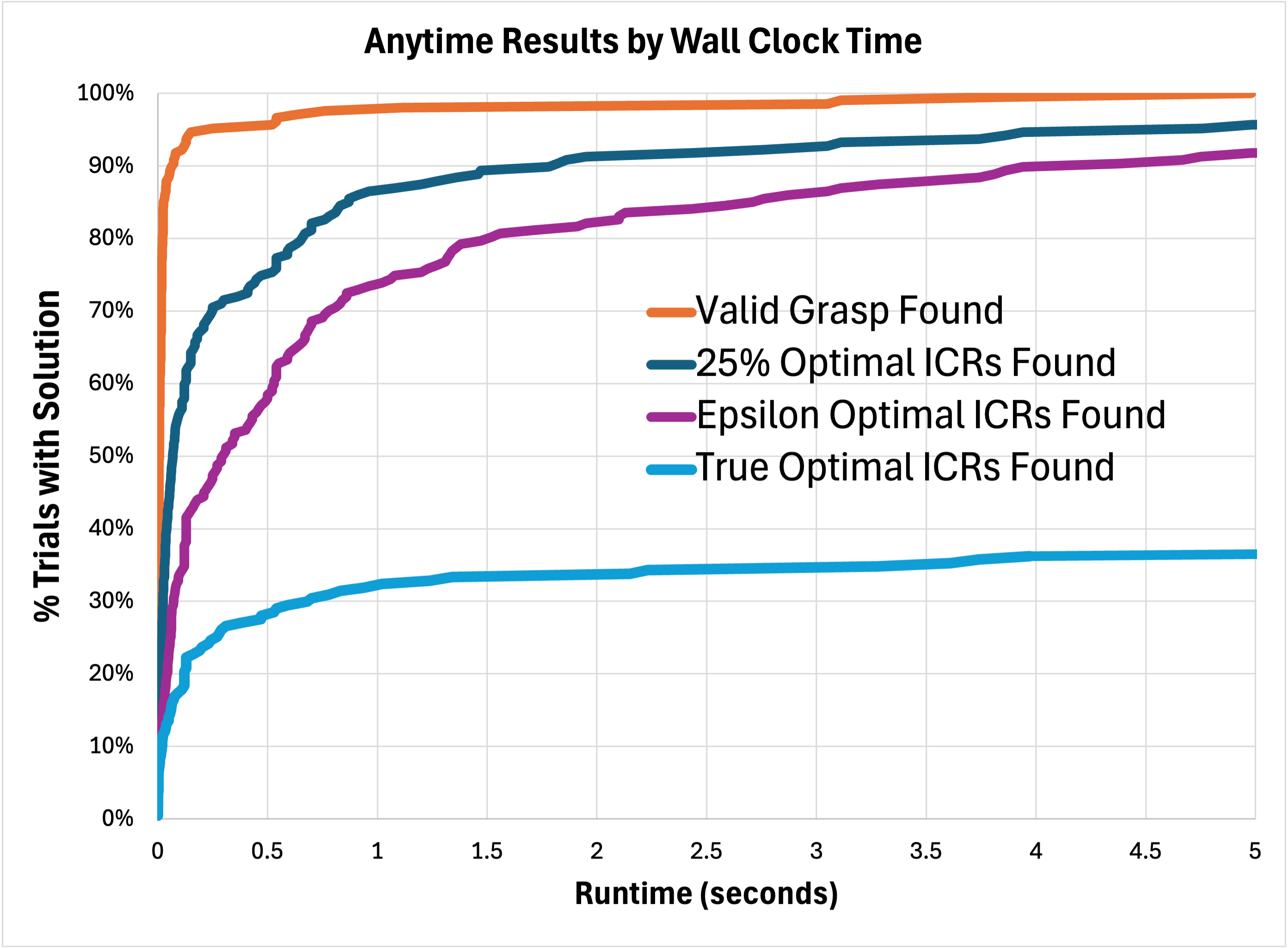}
    \caption{Left to right:   (1) Contact regions Phoka vs. ours are similar; (2) our runtimes (solid) show large improvements over brute force (dashed);  (3) optimal ICRs for simple test objects;  (4) anytime results show near-optimal solutions are returned quickly.}
    \label{fig:runtime}
\end{figure*}

Figure~\ref{fig:runtime} (Left) shows a visual comparison of largest ICRs identified by our algorithm and that of Phoka et al.~\cite{phoka2012optimal}.    Table~\ref{tab:phoka} compares runtimes.  All of our times improved upon the figures reported in~\cite{phoka2012optimal}.  Runtimes less than 500ms are shown in bold.   Note that some grasp was found within 500ms for all examples, and for eight of the twelve examples our algorithm found optimal grasps within that time, although it sometimes took additional time to prove optimality.

\medskip\noindent
{\bf Brute Force Comparison: $L_\infty$-Distance Transforms.}
In spite of interesting research developments in $L_{\infty}$ Voronoi diagrams~\cite{bukenberger2022constructing}, there are no algorithms or libraries available to apply these algorithms in arbitrary dimension and no other competing algorithms for grasps having arbitrary numbers of contacts.   Best practice is therefore a brute force approach.   In this section, we compare the runtime of our algorithm to a brute force approach which first computes the quality metric for every grasp in $C_d$ and then finds the maximal ICR using the $L_\infty$-Distance Transform, for which fast libraries are available.\footnote{For example {\tt scipy.ndimage.distance\_transform\_bf} using the chessboard metric.}
We tested all combinations of:
\begin{itemize}
    \item \textbf{Shape}: Banana, Controller, Ellipse, Boomerang
    \item \textbf{Resolution}: 32, 64, 128, 256, 512
    \item \textbf{Friction Cone}: $10^{\circ}$, $15^{\circ}$, $20^{\circ}$
    \item \textbf{Number of Contacts}: 2, 3, 4, 5
\end{itemize}


\begin{table}[ht]
\centering
\resizebox{0.48\textwidth}{!}{
\begin{tabular}{|l|c|c|c|c|c|c|c|c|c|c|c|}
\hline
{\bf Shape} & {\bf Time} & {\bf Time} & {\bf \#} & {\bf \#grasp} & {\bf \%grasp} & {\bf \#} & {\bf true} & {\bf iter} & {\bf eps} \\
 & {\bf (BF)} & {\bf (Iter)} & {\bf Iter's} & {\bf checks} &  {\bf checks} & {\bf Simp's} & {\bf rad} & {\bf rad} & {\bf ret} \\
\hline
\hline
\multicolumn{10}{|c|}{\textbf{2 Contacts}} \\
\hline
Banana    & {\bf 0.22} & 0.34 & 420 & 454 & 5.6\% & 926 & 1 & 1 & 1 \\
Controller & 0.22 & {\bf 0.14} & 254 & 316 & 3.9\% & 552 & 2 & 2 & 1 \\
Ellipse   & 0.22 & {\bf 0.13} & 245 & 291 & 3.6\% & 527 & 3 & 2 & 1  \\
Prongs    & {\bf 0.22} & 0.59 & 565 & 591 & 7.3\% & 1204 & 1 & 1 & 1  \\
\hline
\hline
\multicolumn{10}{|c|}{\textbf{3 Contacts}} \\
\hline
Banana    & 9.68 & {\bf 0.23} & 106 & 3.96K & 1.2\% & 633 & 8 & 6 & 4.0  \\
Controller & 9.82 & {\bf 0.32} & 199 & 2.23K & 0.7\% & 1215 & 7 & 5 & 3.6  \\
Ellipse   & 9.74 & {\bf 0.20} & 96 & 3.42K & 1.0\% & 578 & 7 & 6 & 4.0  \\
Prongs    & 9.79 & {\bf 0.16} & 98 & 2.49K & 0.7\% & 587 & 7 & 6 & 4.2  \\
\hline
\hline
\multicolumn{10}{|c|}{\textbf{4 Contacts}} \\
\hline
Banana    & 338.17 & {\bf 4.67} & 70 & 128K & 1.2\% & 1222 & 10 & 8 & 8  \\
Controller & 334.37 & {\bf 2.84} & 111 & 75K & 0.7\% & 2024 & 9 & 7 & 7  \\
Ellipse   & 329.47 & {\bf 8.16} & 37 & 235K & 2.2\% & 559 & 14 & 10 & 8  \\
Prongs    & 328.03 & {\bf 3.95} & 85 & 105K & 1.0\% & 1476 & 9 & 8 & 8  \\
\hline
\hline
\multicolumn{10}{|c|}{\textbf{5 Contacts}} \\
\hline
Banana    & OOM & {\bf 58.87} & 131 & 1.52M & 0.6\% & 9.0K & NA & 7 & 8  \\
Controller & OOM & {\bf 56.27} & 124 & 1.44M & 0.5\% & 9.0K & NA & 7 & 8  \\
Ellipse   & OOM & {\bf 171.77} & 175 & 4.49M & 1.7\% & 13.5K & NA & 10 & 4  \\
Prongs    & OOM & {\bf 27.27} & 306 & 0.57M & 0.2\% & 25.7K & NA & 6 & 7  \\
\hline
\end{tabular}
}
\caption{Statistics include brute force runtime in seconds (Time BF), iterative runtime in seconds (Time Iter), number of iterations ($\#$Iter's), number and percentage of grasps checked ($\#$grasp checks and $\%$grasp checks), number of simplices ($\#$Simp's), radius of the largest ICR (true rad), the radius of the ICR returned by the iterative algorithm (iter rad), the possible error in radius returned by the iterative algorithm (eps).  Shape resolutions are 128 with a $15^{\circ}$ friction cone. }
\label{tab:allStats}
\end{table}

Run times are shown in Figure~\ref{fig:runtime} (Left plot) and are given in log-scale.  Run times for the iterative algorithm are shown as solid lines, and run-times for brute force are shown as dashed lines.   Brute force results for 5 contacts beyond 64 samples and 4 contacts beyond 256 samples could not be computed due to ``out of memory'' errors.  Best grasps for the iterative algorithm for trials having resolution of 128 and contact friction cone angle of $15^{\circ}$ are shown graphically in Figure~\ref{fig:runtime} (Middle).    
Figure~\ref{fig:bunny_fc_dt} (Right) shows optimal ICRs identified by our algorithm for grasps of the bunny having various number of contacts up to 7.  To our knowledge, the results shown in this paper are the first $\epsilon$-optimal ICRs to be computed for general grasps having 4 or more contacts. 
Table~\ref{tab:allStats} gives detailed statistics.

\medskip\noindent
{\bf Anytime Results.}
Figure~\ref{fig:runtime} (Right) illustrates the anytime nature of the algorithm.   To collect these statistics, we examined all trials in which the brute force algorithm ran to completion, such that the True Optimal solution was known.   This set consisted of 207 trials containing the 9 objects shown in this paper and friction cone angles of $10^\circ$, $15^\circ$, and $20^\circ$.   The breakdown in terms of contacts was 2-contacts (35\%), 3-contacts (29\%), 4-contacts (24\%), and 5-contacts (12\%).

Valid grasps are typically found very early in a trial, with 80\% of trials having found a valid grasp within 5\% of total trial time.   Grasps that are $\epsilon$-optimal are found at various points during the trial time, although it then can take more time to prove that they are optimal.   The algorithm was implemented with early exit, and many trials never identify the true optimal ICR's.   However, more than 40\% of them do find the true optimal ICR.  For half of those trials, the true optimal ICR is the last hypercube that is grown and enables the trial to terminate.   Considering wall clock time, which is shown in the Figure, at just 250ms, more than 95\% of trials have found a valid grasp, more than 70\% of trials have ICRs within 25\% of optimal, and nearly half of all trials have already found an $\epsilon$-optimal grasp, with half of those being the true optimal.    By 5 seconds, more than 90\% of trials have $\epsilon$-optimal grasps and all have some grasp that is valid.

\begin{figure}[h!]
  \centering
    \includegraphics[width=0.95\columnwidth]{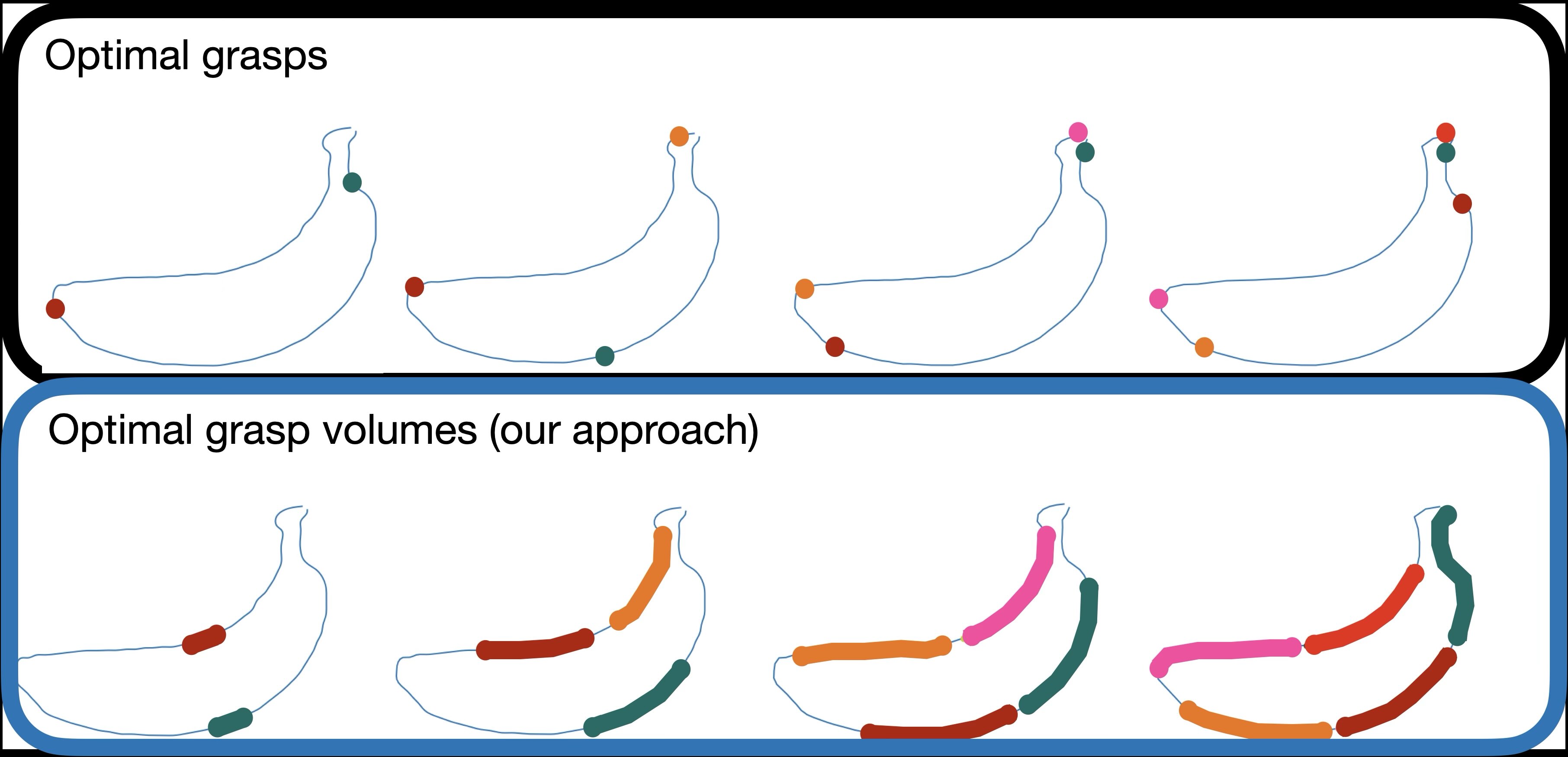}
    \vspace*{-0.1in}
    \caption{(Top) Optimal grasps found using the point based metric in~\cite{ferrari1992planning}.  (Bottom) Optimal ICRs identified using the identical metric to classify valid vs. invalid grasps.}
    \label{fig:otherMetrics}
\end{figure}

\medskip\noindent
{\bf Point vs. Volume Metric Qualitative Comparison.}
We can qualitatively compare ICRs, which are a volume based quality metric, against traditional point based quality metrics.   Figure~\ref{fig:otherMetrics} shows a qualitative comparison.   To obtain this result, globally optimal grasps were first computed using a brute force approach for the Ferrari and Canny wrench space ball quality metric~\cite{ferrari1992planning}.   This is a highly cited and widely used general purpose quality metric for grasp and manipulation planning.   The grasps shown have the globally optimal quality value based on this metric.   Then, we ran our algorithm to identify optimal ICRs using {\it exactly the same metric} with a threshold $q>0$ to classify grasps as valid vs. invalid.   The ICRs returned are the $\epsilon$-optimal independent contact regions, i.e., the largest axis aligned volumes for which the given metric produces a valid grasp.  We can consider ICRs found using the Ferrari and Canny metric as a volumetric measure based on that metric.  We observe that this volumetric measure identified contacts qualitatively more likely to lead to a successful outcome given real-world constraints, dynamics, and uncertainties.

\medskip\noindent
{\bf Robustness to Uncertainties.}
To test robustness to uncertainties, a simple control policy was implemented to move towards ICRs and apply balanced forces using impedance control.   Experiments are ongoing, but the policy has been found to be robust to $\pm20\%$ size changes, $1.8cm$ position variation, and substantial geometric differences (Figure~\ref{fig:teaser}).

\section{Discussion}

As illustrated in Figure~\ref{fig:runtime} (Left plot), the iterative algorithm presented in this paper outperforms brute-force for all but the very lowest resolution and 2-contact cases, with the improvement in performance becoming substantial ($\approx$100X) as number of contacts increases.  This improvement makes computation of ICRs feasible for higher contact grasps for the first time. 

The proportion of run time spent on major subroutines is dominated by finding the best candidate and growing hypercubes, with the latter being most dominant for higher resolution or higher numbers of contacts.   Dependence on resolution comes from the fact that the hypercube size grows with resolution to the $dth$ power, with $d$ the number of contacts.   Considering only the time required to grow the largest hypercube that enables termination, runtime requirements for hypercube growing, expressing region size as a fraction $\alpha$ of the number of samples $p$ on a shape is $O((\alpha p)^d)$.   Complexity of the iterative Delaunay Triangulation depends on the number of invalid grasps added $n_b$, and can be expressed as $O(n_b^{\lfloor\frac{d}{2}\rfloor})$.   Both are exponential in $d$.  However, higher dimensional triangulations tend to require relatively few insertions of invalid points and few simplices, and time required to grow hypercubes becomes the practical limiting factor in the algorithm's performance.

The linear growth in runtime in the log plot in Figure~\ref{fig:runtime} (Left plot) reflects exponential growth with sample size.  However, the iterative algorithm is practically much faster, as only most promising hypercubes need to be checked.  In this approach, runtime speedup is limited by the fractional volume of the Order Simplex filled by the largest hypercube, and one promising future area of research is to address this limitation either with an analytical approach or sampling based techniques that would avoid exhaustive sampling of the largest hypercube.

We expected increased friction cone angles to result in longer runtimes, as they might be expected to produce larger contact regions, and thus larger hypercubes to grow.   This effect is seen in the runtimes for Phoka and colleagues~\cite{phoka2012optimal} (Table~\ref{tab:phoka}).  However, we did not observe this effect in our results.  No effect was observed for $\mu$ on runtimes ($P > 0.6$).  A slight effect was observed for $\mu^d$ on runtimes ($R^2 = 0.02, P < 0.05$).  However, the effect was in the opposite direction expected, {\it with increasing friction cones producing faster runtimes.}   It is possible this effect was due to larger friction cones making it easier to find valid solutions which enabled the algorithm to converge faster.   This possibility is supported by larger $\mu$ producing fewer iterations ($P < 0.05$).

A similar phenomenon may be responsible for the dip in runtimes when moving from a resolution of 32 to a resolution of 64.  With 32 samples, the algorithm often ends up subdividing the entire space to prove that it has found the best region, making it a brute force algorithm with overhead.   However, when moving to 64 samples, larger regions exist and are found early in the search, eliminating this problem.

\medskip\noindent{\bf Limitations}
One limitation of the algorithm is that the result returned is guaranteed to be optimal only within an $\epsilon$ bound.   If a true global optimal solution is desired, additional search must be performed within previously explored simplices.   Implementing such a search is a topic for future work.

A second limitation is the use of an occlusion point to break up the surface. If that occlusion point is chosen poorly, the true optimal grasp may not be found. The initial simplex may be expanded to allow the curve to wrap past the occlusion point and avoid this problem (e.g., as in~\cite{caroli2016delaunay}). 
 
Finally, the algorithm currently is restricted to 2D objects with contact on the perimeter.  It can be applied to thin objects, extruded objects, and to arbitrary 3D object slices (with some care put towards the definition of valid grasp).   A path to lift this restriction is outlined in the Appendix.




\section{Conclusion}

We have presented an iterative algorithm for computing maximal independent contact regions that is fast, complete, works for any number of contacts, and can be used as an anytime algorithm.   The key insight behind this approach is that Delaunay triangulation, which extends well to higher dimensions, offers a powerful way to search the grasp configuration space for regions where large ICRs may be found.  The resulting ICRs can be used as guidance for real-time grasp planning, learning, and shared control.   We look forward to exploring these and many other applications.

 \section{Appendix - A Path to General 3D Grasps}

 A sketch of a general 3D solution follows.  Any manifold mesh can be flattened in a manner that best preserves properties of interest, such as local surface areas (e.g.,~\cite{sawhney2017boundary}).  ICRs on the surface of the mesh will appear as sets of circles on these flattened meshes, such that any grasp composed of one contact within each circle can be a valid grasp.   The algorithm described in this manuscript can be used to identify such regions.   For three contacts, for example, use the described algorithm to search for the largest empty circumsphere in a six-dimensional space, where the 6D space consists of three copies of the flattened mesh.  Instead of growing hypercubes, grow circles outward for each flattened mesh, testing all combinations of grasps represented by points in the grown 6D volume.   Adjust epsilon bounds to account for the change in region geometry.   We are excited to explore this and other extensions such as soft area contacts, ICRs for rolling and sliding, and ICRs for policy learning in future research.

\bibliographystyle{IEEEtranS}
\bibliography{references}

\end{document}

%% file: macros_packages/math.tex
\DeclareMathSymbol{\shortminus}{\mathbin}{AMSa}{"39}

\newlength{\myleftlen}

\newlength{\myrightlen}

%% file: macros_packages/manip.tex
\newcommand*{\point}[1][]{\ensuremath{\bm{p}\ifx\\#1\\\else_{#1}\fi}}
\newcommand*{\normalvec}[1][]{\ensuremath{\bm{n}\ifx\\#1\\\else_{#1}\fi}}
\newcommand*{\tangentvec}[1][]{\ensuremath{\bm{t}\ifx\\#1\\\else_{#1}\fi}}

\newcommand*{\force}[1][]{\ensuremath{\bm{f}\ifx\\#1\\\else_{#1}\fi}}
\newcommand*{\torque}[1][]{\ensuremath{\tau\ifx\\#1\\\else_{#1}\fi}}    
\newcommand*{\wrench}[1][]{\ensuremath{\bm{w}\ifx\\#1\\\else_{#1}\fi}}

\newcommand*{\grasp}[1][]{\ensuremath{\mathcal{G}\ifx\\#1\\\else_{#1}\fi}}

\newcommand{\ceq}{\ensuremath{\coloneqq}}

%% file: macros_packages/utils.tex
\SetCommentSty{mycommfont}
\SetKwRepeat{Struct}{struct \{}{\}}%

\SetAlgoProcName{Subroutine}{routine}

\makeatletter
\AtBeginEnvironment{procedure}{\let\c@algocf\c@procedure}
\makeatother
\SetAlgoFuncName{Data Structure}{struct}

\makeatletter
\AtBeginEnvironment{function}{\let\c@algocf\c@function}
\makeatother
\makeatletter
\renewcommand{\Indentp}[1]{%
  \advance\leftskip by #1
  \advance\skiptext by -#1
  \advance\skiprule by #1}%
\renewcommand{\Indp}{\algocf@adjustskipindent\Indentp{\algoskipindent}}
\renewcommand{\Indm}{\algocf@adjustskipindent\Indentp{-\algoskipindent}}
\makeatother

\colorlet{cellbg}{black!10}
\newlength\heavyline    
\newlength\lightline    

\definecolor{MoonRaker}{rgb}{0.745,0.78,0.945}
\definecolor{MacaroniandCheese}{rgb}{1,0.756,0.517}
\definecolor{Sulu}{rgb}{0.576,0.933,0.525}
\definecolor{DoveGray}{rgb}{0.435,0.435,0.435}
\definecolor{Portage}{rgb}{0.541,0.603,0.901}
\definecolor{WestSide}{rgb}{1,0.529,0.058}
\definecolor{LaPalma}{rgb}{0.18,0.725,0.101}
\definecolor{Silver}{rgb}{0.752,0.752,0.752}